\documentclass[lettersize,journal]{IEEEtran}
\usepackage{amsmath,amsfonts}
\usepackage{algorithmic}
\usepackage{algorithm}
\usepackage{array}
\usepackage[caption=false,font=normalsize,labelfont=sf,textfont=sf]{subfig}
\usepackage{textcomp}
\usepackage{stfloats}
\usepackage{url}
\usepackage{xcolor}
\usepackage{multirow}
\usepackage{booktabs}
\usepackage{verbatim}
\usepackage{graphicx}
\usepackage{cite}     
\usepackage[dvipsnames]{xcolor}      
\definecolor{CiteBrightGreen}{HTML}{00E676} 

\PassOptionsToPackage{colorlinks,linkcolor=blue,citecolor=blue,urlcolor=blue}{hyperref}
\usepackage{orcidlink}  

\begin{document}

\title{IMAGGarment: Fine-Grained Garment Generation for Controllable Fashion Design}

\author{Fei Shen,
        Jian Yu,
        Cong Wang,
        Xin Jiang,
        Xiaoyu Du,
        and Jinhui Tang,~\IEEEmembership{Senior~Member,~IEEE}
\thanks{Fei Shen is with the School of Computer Science and Engineering, Nanjing University of Science and Technology, Nanjing, 210094, China, and also with the NExT++ Research Centre, National University of Singapore, Singapore, e-mail: {shenfei29@nus.edu.sg}}

\thanks{Jian Yu, Xin Jiang, and Xiaoyu Du are with the School of Computer Science and Engineering, Nanjing University of Science and Technology, Nanjing, 210094, China. e-mail: {jianyu@njust.edu.cn; xinjiang@njust.edu.cn; duxy@njust.edu.cn}.}

\thanks{Cong Wang is with the State Key Laboratory for Novel Software Technology and the School of Computer Science, Nanjing University, Nanjing, 210023, China. e-mail: {cw@smail.nju.edu.cn}}

\thanks{Jinhui Tang is with the School of Computer Science and Engineering, Nanjing University of Science and Technology, Nanjing, 210094, China, and also with the College of Information Science and Technology and
Artificial Intelligence, Nanjing Forestry University, Nanjing 210037, China, e-mail: jinhuitang@njust.edu.cn. (Corresponding author: Jinhui Tang.)}}

\markboth{Journal of \LaTeX\ Class Files,~Vol.~14, No.~8, August~2021}%
{Shell \MakeLowercase{\textit{et al.}}: A Sample Article Using IEEEtran.cls for IEEE Journals}


\maketitle

\begin{abstract}
This paper presents IMAGGarment, a fine-grained garment generation (FGG) framework that enables high-fidelity garment synthesis with precise control over silhouette, color, and logo placement. 
Unlike existing methods that are limited to single-condition inputs, IMAGGarment addresses the challenges of multi-conditional controllability in personalized fashion design and digital apparel applications.
Specifically, IMAGGarment employs a two-stage training strategy to separately model global appearance and local details, while enabling unified and controllable generation through end-to-end inference.
In the first stage, we propose a global appearance model that jointly encodes silhouette and color using a mixed attention module and a color adapter.
In the second stage, we present a local enhancement model with an adaptive appearance-aware module to inject user-defined logos and spatial constraints, enabling accurate placement and visual consistency.
To support this task, we release GarmentBench, a large-scale dataset comprising over 180K garment samples paired with multi-level design conditions, including sketches, color references, logo placements, and textual prompts. 
Extensive experiments demonstrate that our method outperforms existing baselines, achieving superior structural stability, color fidelity, and local controllability performance. Code, models, and datasets are publicly available at \url{https://github.com/muzishen/IMAGGarment}.
\end{abstract}


\section{Introduction}
Fine-Grained garment generation (FGG) aims to synthesize high-quality garments with precise control over garment silhouette, color scheme, logo content, and spatial placement. As personalized fashion and the digital apparel market grow rapidly, fine-grained controllability~\cite{zhang2025diffusion, li2022learning, zhang2025viton,zhou2023lc} is increasingly crucial for applications in fashion design and e-commerce.

In traditional garment ideation~\cite{chowdhury2022garment,jin2024human} and visualization~\cite{durupynar2007virtual,pa2025smart}, designers analyze line drawings to establish silhouette and construction, then select color palettes and materials, and finally arrange brand elements such as logos and trims. This manual workflow has two persistent drawbacks. First, it is time consuming: to match the specification, edits must be applied object by object and view by view; in a seasonal collection, even identical panels within the same board are recolored or relabeled one at a time, which does not scale. Second, it is error prone and inconsistent: small deviations in hue, shading, or logo placement arise across artists and rounds of revision, yielding mismatches across styles, sizes, and camera viewpoints. As project scope grows, these issues inflate turnaround time and complicate quality control and version management.

\begin{figure*}
  \vspace{-0.4cm}
  \includegraphics[width=0.95\textwidth]{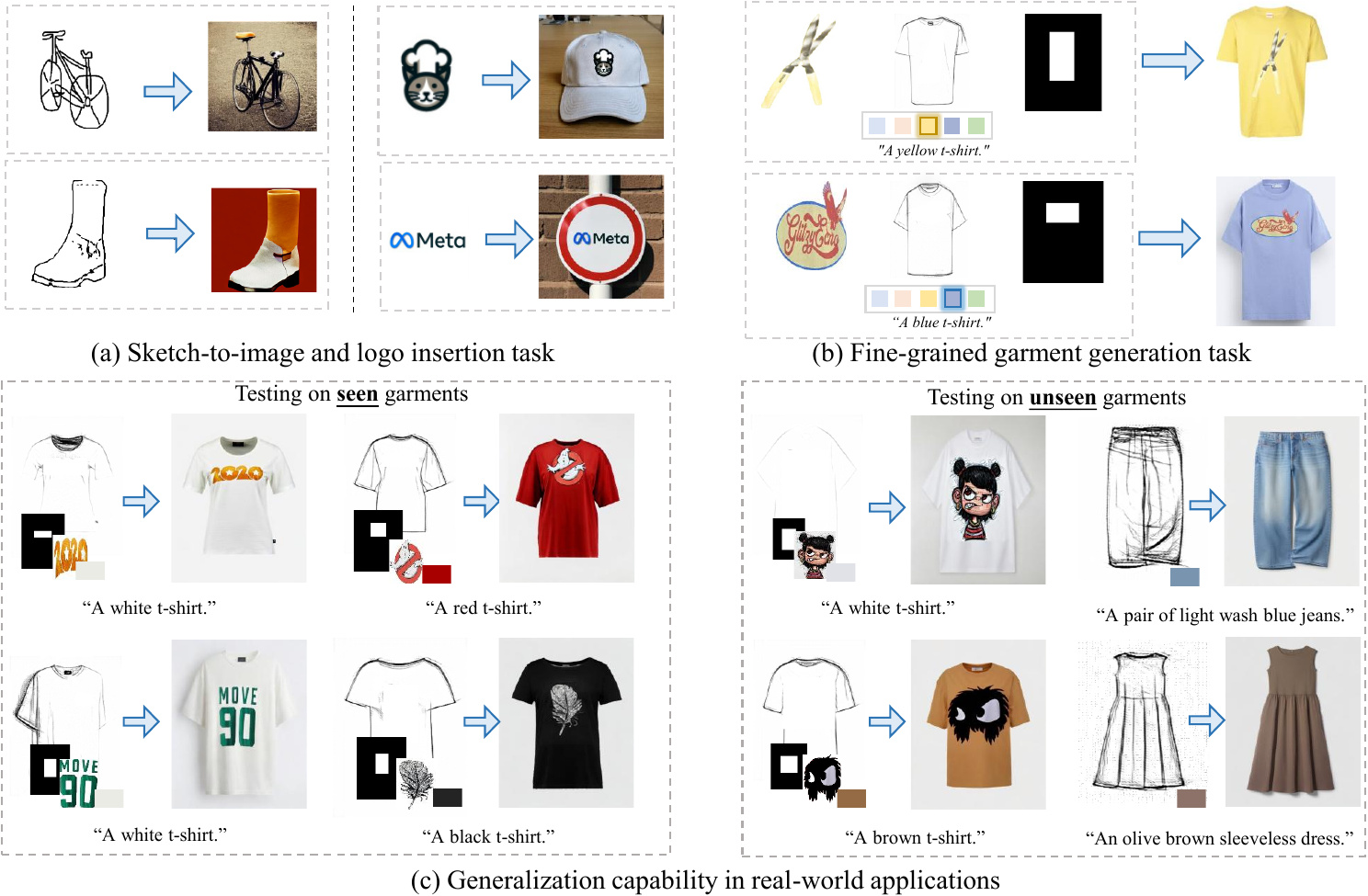}
  \centering
  \caption{
\textbf{Comparison of (a) existing sketch-to-image and logo insertion tasks with (b) our proposed fine-grained garment generation (FGG) task,} which enables precise and controllable synthesis of garment structure, color, logo, and spatial placement. Unlike previous tasks that rely on a single input condition, FGG is tailored for real-world fashion design workflows by integrating multiple conditional controls. 
  }
  \label{fig:example}
\end{figure*}

Recently, image synthesis~\cite{podell2023sdxl, lipman2022flow,su2022drawinginstyles,chen2025human} has made notable progress in tasks such as sketch-to-image generation~\cite{voynov2023sketch, isola2017image,koley2024s,koley2024text} and logo insertion~\cite{kim2024conditional,zhang2024anylogo,zhu2024logosticker} (as illustrated in Fig.~\ref{fig:example} (a)), demonstrating basic capabilities in structural and content-level control.
However, these tasks~\cite{voynov2023sketch,kim2024conditional,yu2025fashiondpo} provide only coarse guidance and rely on single-condition inputs (\emph{e.g.}, sketch or color), lacking the fine-grained controllability needed to model the nuanced interactions between global structure and local details in garment design.
Although sequential or modular combinations may offer partial solutions, they~\cite{ye2023ip, zhang2023adding, rombach2022high} fail to explicitly disentangle and jointly model global attributes (\emph{e.g.}, silhouette, color) and local appearance details (\emph{e.g.}, logo content and spatial placement).
Without unified control mechanisms, these approaches~\cite{ye2023ip, zhang2023adding, rombach2022high} often suffer from condition entanglement, conflicting objectives, and visual inconsistencies, ultimately falling short of the high standards required in real-world fashion design.
In contrast, practical fashion design~\cite{chowdhury2022garment,jin2024human} requires joint control over multiple interdependent factors: designers determine global attributes such as silhouette and color, followed by fine-tuning of local elements like logos and their placement.
To support this process, a unified generation task that clearly separates and coordinates global and local attributes is essential for controllable and high-fidelity synthesis.

To address these limitations, we propose a new task: fine-grained garment generation (FGG), as illustrated in Fig.~\ref{fig:example} (b).
FGG is formulated as a unified multi-conditional garment synthesis task, taking a textual prompt, garment silhouette, color palette, and spatially constrained logos as joint inputs. It aims to generate garments that faithfully reflect high-level structural intent and fine-grained local styling cues.
FGG is specifically designed to mirror real-world fashion workflows, where designers must coordinate diverse input modalities to express creative intent. Unlike conventional approaches that process each condition independently or sequentially, FGG emphasizes joint modeling and hierarchical reasoning across input types.
It goes beyond simple task combinations by enforcing consistent integration of global and local attributes within a unified generation framework, enabling nuanced control over the overall structure and detailed appearance.
Specifically, \textbf{FGG task introduces three key challenges:}
(1) maintaining visual and semantic consistency across heterogeneous input conditions,
(2) resolving conflicts between global structures and localized visual elements, and
(3) generalizing to unseen condition combinations without retraining (see Fig.~\ref{fig:example}(c)).
FGG thus marks a fundamental shift from single-condition or loosely coupled pipelines toward a unified, design-intent-driven generation paradigm that better reflects the complexity of real-world garment design.

To this end, we propose IMAGGarment, a two-stage training and end-to-end inference framework tailored for fine-grained garment generation.
Unlike prior methods that rely on single-condition inputs or simple feature fusion, our framework is explicitly designed to achieve fine-grained controllability under multiple, interdependent constraints.
In the first stage, we propose a global appearance model with a mixed attention module and a color adapter to jointly encode garment silhouette and color palette, improving overall appearance fidelity and mitigating condition entanglement.
In the second stage, we present a local enhancement model equipped with an adaptive appearance-aware module to inject user-defined logos and their spatial constraints, enabling precise logo placement while preserving global consistency.
To further promote research in this direction, we release GarmentBench, a large-scale dataset comprising over 180k garment samples annotated with rich multi-level design conditions, including silhouette sketches, color references, logo placements, and textual prompts.
Extensive experiments demonstrate that IMAGGarment significantly outperforms existing baselines in terms of structural stability and local controllability.
To summarize, the main contributions are listed as follows:
\begin{itemize} 
\item We propose {IMAGGarment}, a controllable garment generation framework that enables precise control over garment structure, color, and logo placement, addressing the challenges of FGG. 
\item We design a {mixed attention module}, {color adapter}, and {adaptive appearance-aware module} to disentangle global structure from local attributes, achieving fine-grained visual control and accurate spatial control. 
\item We release {GarmentBench}, a large-scale dataset with diverse garments and rich multi-conditional annotations, serving as a valuable benchmark for controllable garment generation research. 
\end{itemize}

The remainder of this paper is organized as follows. Section~\ref{sec:rw} surveys prior work on garment generation, encompassing GAN-based techniques and diffusion-based controllable generation. Section~\ref{sec:method} describes the proposed IMAGGarment methodology, comprising a global appearance model with mixed attention and a color adapter, a local enhancement model with the A3 module, and the associated training and inference strategies. Section~\ref{sec:exp} presents the experimental protocol and results, including the GarmentBench dataset and evaluation metrics, implementation details, and results and analysis. Section~\ref{sec:conclusion} concludes the paper.

\section{Related Work}\label{sec:rw}

\subsection{GAN-Based Methods}

Early approaches~\cite{10.1145/3386569.3392386,10.1145/3450626.3459760,wu2022deepportraitdrawinggeneratinghumanbody,ghosh2019interactive,chen2018sketchygan,li2020staged} to garment generation predominantly build on generative adversarial networks (GANs)\cite{creswell2018generative,mirza2014conditional,men2020controllable}, with a major line devoted to sketch-to-image translation\cite{liu2017auto} that learns spatial mappings from structural cues. Representative systems such as DeepFaceDrawing~\cite{10.1145/3386569.3392386} and DeepFaceEditing~\cite{10.1145/3450626.3459760} decompose sketches into semantic components and progressively assemble photorealistic results, while DeepPortraitDrawing~\cite{wu2022deepportraitdrawinggeneratinghumanbody} extends this paradigm to full-body synthesis via local-to-global pipelines. Interactive frameworks~\cite{ghosh2019interactive} further introduce gating mechanisms for user-guided editing, and DALColor~\cite{10.1145/3240508.3240661} combines WGAN-GP~\cite{gulrajani2017improved} with line-art colorization for refined appearance control. Beyond sketches, related GAN-based efforts explore pose- or part-guided generation~\cite{ma2017pose,siarohin2018deformable}, leveraging learned warping or deformable alignment to better propagate structural constraints from sources to targets.

However, these methods~\cite{10.1145/3386569.3392386,10.1145/3450626.3459760,wu2022deepportraitdrawinggeneratinghumanbody,ghosh2019interactive} are largely restricted to single-condition settings (e.g., sketches or poses alone), making it difficult to support real-world fashion scenarios that require joint control over multiple factors such as silhouette, garment layers, color/pattern, and local embellishments. Moreover, adversarial training is prone to instability and visual artifacts~\cite{ma2017pose,men2020controllable,siarohin2018deformable}, and the reliance on paired or carefully aligned supervision limits robustness to occlusion, diverse body shapes, and open-world catalogs. As a result, while GAN-based pipelines can produce plausible textures under constrained conditions, they struggle to achieve reliable, fine-grained, and multi-conditional controllability at scale.

\subsection{Diffusion-Based Methods}
Diffusion models~\cite{ho2020denoising, song2020score, gao2025faceshot} have achieved strong progress in conditional image generation owing to their iterative denoising process and flexible conditioning interfaces. To improve controllability with minimal modification to large backbones, plugin-based approaches such as IP-Adapter~\cite{ye2023ip}, ControlNet~\cite{zhang2023adding}, and BLIP-Diffusion~\cite{li2023blip} inject external conditions (e.g., reference images, structural maps, or language cues) through lightweight adapters. In parallel, reference-guided or dual-stream designs~\cite{shen2024imagdressing, lin2024dreamfit, chen2024magic, xu2024ootdiffusion} propagate features from exemplars alongside text/image prompts, thereby strengthening identity preservation and fine control during sampling.

In fashion-related applications, DiffCloth~\cite{zhang2023diffcloth} supports localized garment edits via part-specific textual prompts, enabling independent control over regions such as sleeves and collars. For logo-centric generation, AnyLogo~\cite{zhang2024anylogo} adopts a dual-state denoising strategy to retain subtle logo details; LogoSticker~\cite{zhu2024logosticker} performs token-based injection to flexibly place logo elements; and RefDiffuser~\cite{kim2024conditional} leverages expert-driven plugins to enhance texture fidelity and spatial alignment. Despite these advances, most methods emphasize either global appearance control or localized editing in isolation. A unified framework that jointly models multiple design conditions, \emph{e.g.}, silhouette and layer topology together with color/pattern and local embellishments, while maintaining structural coherence across the denoising trajectory remains underexplored. 

\begin{table}[t]
    \renewcommand{\arraystretch}{1.0}
    \centering
    \caption{Definitions of main symbols used in this paper.}
    \begin{tabular}{l|l}
        \hline
        Notation & Definition \\ \hline
        ${t}$ & Timestep\\
        ${Z_t}$ & Latent feature at $t$ step\\
        ${Z_m}$ & Output of mixed attention\\
        ${x_0}$ & Real image\\
        ${x_t}$ & Noisy data at $t$ step\\
        ${G}$ &  Garment image\\
        ${L}$ & Logo image\\
        ${M}$ & Mask image\\
        ${C_g}$ & Feature of garment image\\
        ${C_l}$ & Feature of logo image\\
        ${C_m}$ & Feature of mask image\\
        ${C_s}$ & Feature of silhouette image\\
        ${C_c}$ & Feature of color image\\
        ${C_t}$ & Feature of text prompt\\
        ${\theta_g}$ & Global appearance model\\
        ${\theta_l}$ & Local enhancement model\\
        ${\epsilon}$  & Gaussian noise\\
        ${\bar{\alpha}_t}$  & Cumulative product of noise weights\\
        ${w}$ & Guidance scale\\
        ${\alpha}$ & Silhouette scale\\
        ${\beta}$ & Color scale\\
        \hline
    \end{tabular}
    \label{tab:notations}
\end{table}

\section{Methodology}\label{sec:method}

\noindent\textbf{Symbol Definition.} To introduce our IMAGGarment method more clearly, we define the main symbols used throughout the paper in TABLE~\ref{tab:notations}.

\noindent\textbf{Task Definition.}
Given a garment silhouette, color palette, user-defined logo, location and an optional text description, fine-grained garment generation (FGG) aims to synthesize high-fidelity garment images with precise control over both global structure and local visual attributes.
The key challenges lie in jointly modeling multi-conditional inputs, maintaining semantic and visual consistency across different design factors, and supporting controllable placement of fine-grained elements such as logos and color regions.

\subsection{Overall Framework}
\begin{figure}[t]
  \centering
  \includegraphics[width=1.0\linewidth]{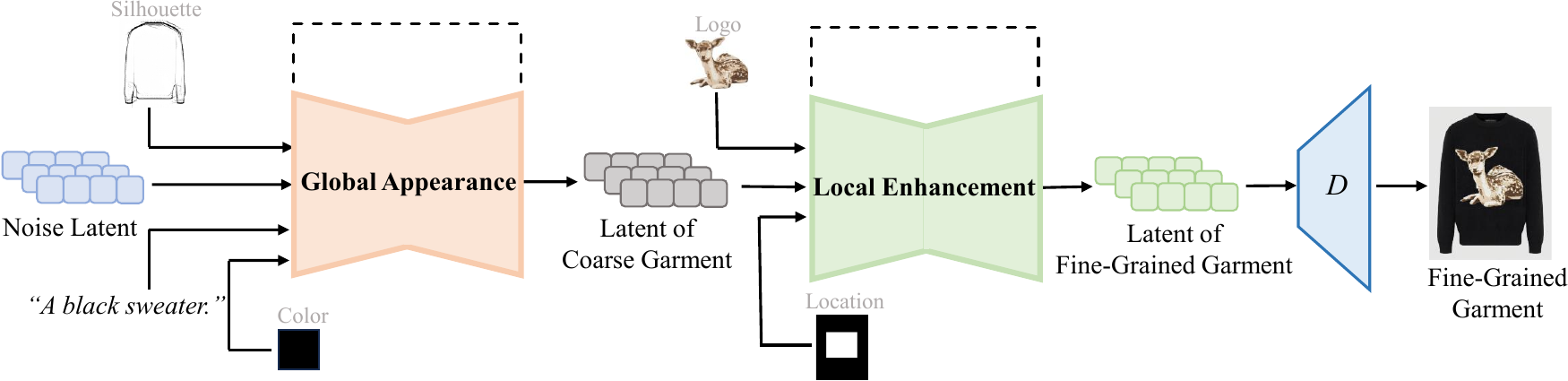}
  \caption{\textbf{Visualization of the IMAGGarment inference pipeline.} 
  The global appearance model generates coarse latent from textual prompts, silhouettes, and colors. The local enhancement model then injects user-defined logos and spatial location constraints to produce the fine-grained garment.}
  \label{fig:pipeline}
\end{figure}

To address the above challenges, we propose IMAGGarment, a conditional diffusion framework tailored for fine-grained garment generation. 
Our framework comprises two components: a global appearance model (stage I) and a local enhancement model (stage II), which explicitly disentangle and jointly control the global appearance and local details under multi-conditional guidance, enabling accurate synthesis of garment silhouette, color, and logo placement.
As illustrated in Fig.~\ref{fig:pipeline}, the global appearance model first generates a latent of coarse garment image conditioned on the textual prompt, garment silhouette, and color palette. 
Subsequently, the local enhancement model refines this latent representation by integrating user-defined logo and spatial constraint, producing the final high-fidelity garment image with fine-grained controllability.
Specifically, the global appearance model (Section~\ref{sec:stage1}) leverages our proposed mixed attention module and color adapter to effectively capture global appearance features from textual descriptions, silhouettes, and colors, while mitigating entanglement among these conditions. 
The local enhancement model (Section~\ref{sec:stage2}) introduces an adaptive appearance-aware module ($A^{3}$ Module) that injects logo content and spatial location constraint into the latent space, achieving precise logo placement.
Finally, the training and inference strategies used in IMAGGarment are summarized in Section~\ref{sec:train_and_infer}.

\subsection{Stage I: Global Appearance Model} \label{sec:stage1}

\begin{figure*}[t]
  \centering
  \includegraphics[width=\linewidth]{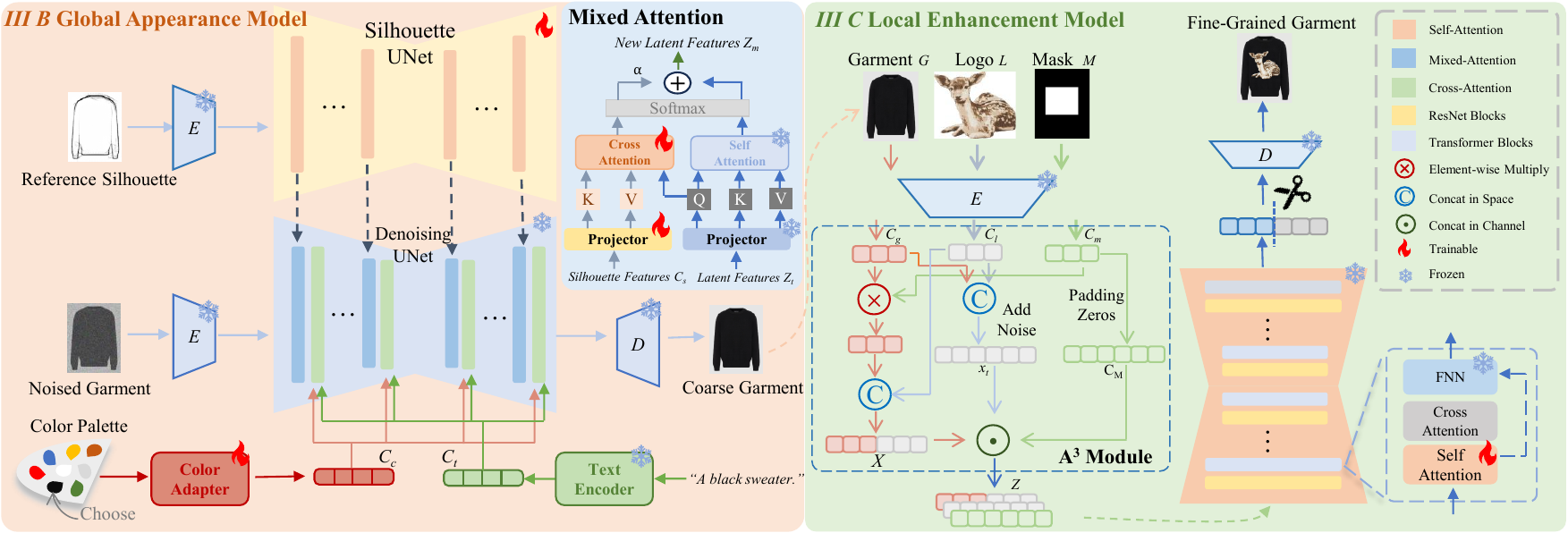}
  \caption{\textbf{Overview of our IMAGGarment framework.} IMAGGarment is a two-stage conditional diffusion framework for fine-grained garment generation. The global appearance model first synthesizes a coarse latent representation from the input text prompt, silhouette, and color palette using a parallel UNet with mixed attention and a color adapter. The local enhancement model then refines this latent by injecting user-defined logos and location constraints through the proposed $A^{3}$ module, enabling precise logo placement and high-fidelity garment generation.}
  \label{fig:architecture}
\end{figure*}

\noindent\textbf{Motivation.} 
Existing garment generation methods~\cite{ye2023ip, zhang2023adding, rombach2022high} typically rely on single-condition inputs (\emph{e.g.}, sketch or text), causing entangled features and limited controllability. To resolve this, we propose a global appearance model that explicitly disentangles silhouette, color, and text, enabling precise multi-conditional control.

\noindent\textbf{Architecture.}
As shown in the left of the  Fig.~\ref{fig:architecture}, our global appearance model comprises two shared frozen VAE encoders, one frozen VAE decoder, a trainable silhouette UNet, a frozen text encoder, a trainable color adapter, and a denoising UNet with the proposed  mixed attention.
Specifically, we first utilize the frozen VAE encoder to project the input reference silhouette into the latent space. Subsequently, we employ a trainable silhouette UNet (structurally identical to the denoising UNet but without cross attention) to extract fine-grained silhouette features, which are then integrated into the frozen denoising UNet via our proposed mixed attention module. Meanwhile, textual features obtained from the frozen CLIP text encoder and color features extracted by the proposed color adapter are further fused into the denoising UNet through cross attention. After multiple denoising iterations, the model generates coarse garment images that precisely align with the reference silhouette and faithfully reflect user-specified color.

\noindent\textbf{Mixed Attention.}
To effectively incorporate reference silhouette features into the denoising UNet without compromising the generative capability of the original UNet, we propose a mixed attention module. 
As shown in Fig.~\ref{fig:architecture}, we extend all self attention layers in the denoising UNet to the proposed mixed attention, which introduces two additional learnable projection layers to align the silhouette features $C_{s}$ with the latent features $Z_{t}$. Formally, the mixed attention is defined as:
\begin{equation}\label{eq:silhouette}
Z_{m} = \text{Softmax}\left(\frac{QK^{T}}{\sqrt{d}}\right)V +  \alpha \cdot \text{Softmax}\left(\frac{Q(K')^{T}}{\sqrt{d}}\right)V',
\end{equation}
where $\alpha$ is a hyperparameter controlling the strength of silhouette conditioning. The projections are computed as follows:
{\small
\begin{equation}\label{eq:cross}
Q=Z_tW_q,\ K=Z_tW_k,\ V=Z_tW_v,\ K'=C_sW'_k,\ V'=C_sW'_v
\end{equation}
}
where $W_{q}, W_{k}, W_{v}$ are frozen parameters of linear projection layers, whereas $W'_{k}, W'_{v}$ are newly introduced learnable parameters of projection layers initialized from $W_{k}$ and $W_{v}$, respectively. 
Our mixed attention facilitates the seamless integration of silhouette features into the denoising UNet, thus ensuring that generated garments maintain precise spatial alignment with the reference silhouette.

\noindent\textbf{Color Adapter.}
Accurate color manipulation is essential for generating garments with fine-grained visual details, significantly enhancing visual quality and realism. However, as the base model's textual prompts cannot reliably produce the intended colors, discrepancies often arise between the generated and expected colors. To address this issue, we propose a dedicated color adapter that explicitly treats color as an independent controllable factor. 
Specifically, given a reference color image, we extract color features $C_{c}$ using a frozen CLIP image encoder combined with a trainable linear layer. 
Subsequently, these color features are integrated into the denoising UNet via a cross attention mechanism, jointly with textual features $C_t$ obtained from the frozen CLIP text encoder:
\begin{equation}\label{eq:color}
Z_{new} = \text{Softmax}\left(\frac{QK_{t}^{T}}{\sqrt{d}}\right)V_{t} + \beta \cdot \text{Softmax}\left(\frac{QK_{c}^{T}}{\sqrt{d}}\right)V_{c},
\end{equation}
where $Q=Z_{t}W_q$, $K_t=C_tW^t_k$, $V_t=C_tW^t_v$, and $K_c=C_cW^c_k$, $V_c=C_cW^c_v$. Here, $W^t_k, W^t_v$ denote frozen parameters of the original cross attention layers in the denoising UNet, while $W^c_k, W^c_v$ are newly introduced trainable projection layers. 
The hyperparameter $\beta$ modulates the adapter’s influence, ensuring precise alignment between generated colors and user specifications.

\subsection{Stage II: Local Enhancement Model}\label{sec:stage2}

\noindent\textbf{Motivation.} 
 Existing methods~\cite{zhang2024anylogo, zhu2024logosticker} typically neglect detailed logo integration or treat it as a separate task, causing poor spatial alignment and visual inconsistency. 
 To address this limitation, we propose a local enhancement model equipped with an adaptive appearance-aware ($A^{3}$) module, explicitly injecting user-defined logos and spatial constraints into the latent space. 
 This design enables precise, consistent control over localized garment details, significantly enhancing visual fidelity.

\noindent\textbf{Architecture.}
As illustrated on the right of Fig.~\ref{fig:architecture}, the local enhancement model comprises a frozen VAE encoder and decoder, a denoising UNet, and an adaptive appearance-aware module ($A^{3}$ module). 
The $A^{3}$ module fuses local conditions, such as logos and spatial constraints, by concatenating them along spatial or channel dimensions, enabling precise control over fine-grained visual elements. 
Given a garment, logo, and placement mask, the model adaptively adjusts the logo's size and position while preserving its visual fidelity. 
To reduce redundancy and focus on local detail refinement, we optimize only the self attention layers of the denoising UNet and discard all cross attention layers, as the global appearance model has already encoded the textual information.

\begin{figure*}[t]
  \centering
\includegraphics[width=0.95\linewidth]{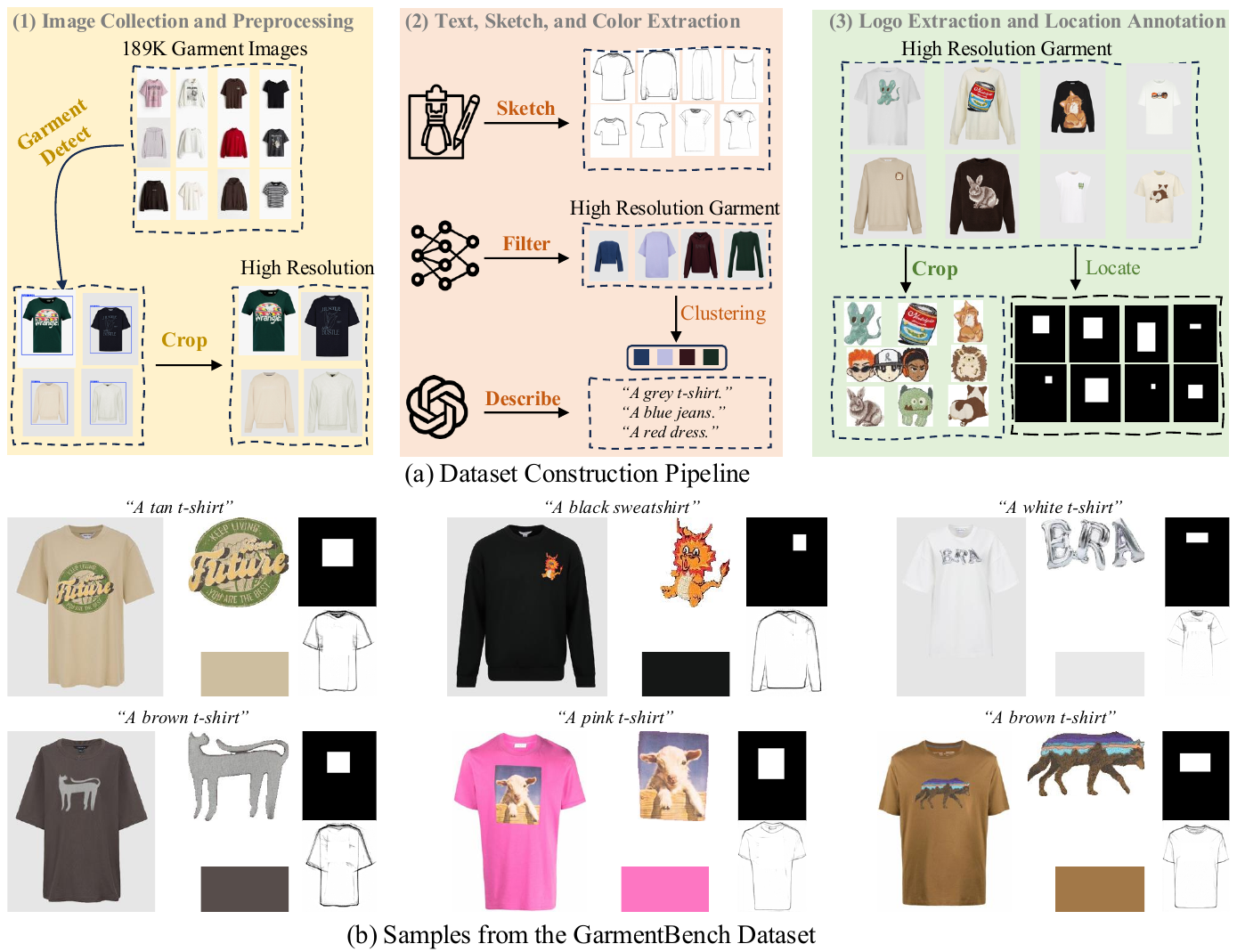}
  \caption{\textbf{Overview of GarmentBench dataset construction pipeline and samples.} (a) Data construction pipeline for GarmentBench. 
(b) Example samples with multimodal annotations: silhouette, logo, text, logo location, and color.}

  \label{fig:dataset_pipeline}
\end{figure*}

\noindent\textbf{\boldmath$A^3$ Module.} 
To precisely integrate fine-grained logo details into designated garment regions, we introduce the adaptive appearance-aware ($A^{3}$) module. 
By fusing image-based conditions across specific dimensions, our $A^{3}$ module enables precise and consistent logo integration.
Specifically, given a coarse garment image $G$, a logo image $L$, and a binary placement mask $M$, we first encode them using a frozen VAE encoder to obtain their corresponding latent features: $C_g \in \mathbb{R}^{4 \times \frac{H}{8} \times \frac{W}{8}}$ and $C_l \in \mathbb{R}^{4 \times \frac{H}{8} \times \frac{W}{8}}$. The mask $M$ is resized via nearest-neighbor interpolation to match the latent resolution, resulting in $C_m \in \mathbb{R}^{1 \times \frac{H}{8} \times \frac{W}{8}}$. We then construct the spatially aligned conditional input as:
\begin{equation}
    X = \text{Concat}(C_g \otimes C_m,\ C_l),  \quad X \in \mathbb{R}^{4 \times \frac{H}{8} \times \frac{W}{4}},
\end{equation}
where $\otimes$ denotes element-wise multiplication and \text{Concat} indicates spatial concatenation along the width dimension. 
To align with $X$, the resized mask $C_m$ is zero-padded to obtain $C_M \in \mathbb{R}^{1 \times \frac{H}{8} \times \frac{W}{4}}$.
Next, we concatenate the garment and logo features to form a clean latent representation:
\begin{equation}
    x_0 = \text{Concat}(C_g,\ C_l),
\end{equation}
and inject noise consistent with the diffusion process:
\begin{equation}
    x_t = \sqrt{\bar{\alpha}_t} \cdot x_0 + \sqrt{1 - \bar{\alpha}_t} \cdot \epsilon, \quad \epsilon \sim \mathcal{N}(0, \mathbf{I}),
\end{equation}
where $x_0$ denotes the clean latent feature obtained by concatenating garment and logo features, and $x_t \in \mathbb{R}^{4 \times \frac{H}{8} \times \frac{W}{4}}$ is the corresponding noisy latent at diffusion timestep $t$. $\bar{\alpha}_t$ is the cumulative product of the noise schedule coefficients, and $\epsilon$ is the Gaussian noise sampled from $\mathcal{N}(0, \mathbf{I})$.
Finally, the full model input is obtained by concatenating the noisy latent $x_t$, the padded mask $C_M$, and the aligned conditional input $X$ along the channel dimension:
\begin{equation}
    Z = \text{Concat}(x_t,\ C_M,\ X), \quad Z \in \mathbb{R}^{9 \times \frac{H}{8} \times \frac{W}{4}}.
\end{equation}
This channel-wise concatenation allows the model to jointly reason over appearance, spatial constraints, and guidance signals, while maintaining compatibility with the UNet architecture for spatially aware logo synthesis.

\subsection{Training and Inference}\label{sec:train_and_infer}

\noindent\textbf{Training.}  
The training process is divided into two stages, each targeting a specific set of objectives with separate optimization strategies. We first train the global appearance model independently to generate a semantically coherent garment representation conditioned on silhouette and color. After verifying its performance, we freeze it and train the local enhancement model to inject fine-grained logos guided by spatial masks. This sequential training avoids gradient interference between heterogeneous objectives and ensures each module converges toward its task-specific goal. Both stages adopt mean squared error (MSE) loss to supervise the denoising process.

\noindent\textbf{Stage I.}  
The global appearance model $\theta_g$ is trained to synthesize garments that align with the target silhouette and color under textual guidance. 
To preserve the generative capacity of the pretrained denoising UNet, we freeze all parameters except those of the silhouette UNet and the cross-attention projections in the mixed attention module. 
Given silhouette features $C_s$, text embeddings $C_t$, and color features $C_c$, we adopt a decoupled training strategy with $L_{\text{silhouette}}$ and $L_{\text{color}}$ losses:
\begin{equation}
\begin{aligned}
    L_{\text{silhouette}} &= \mathbb{E}_{x_0, \epsilon, C_t, C_s, t} \left\| \epsilon - \epsilon_{\theta_g}(x_t, C_t, C_s, t) \right\|^2, \\
    L_{\text{color}} &= \mathbb{E}_{x_0, \epsilon, C_t, C_c, t} \left\| \epsilon - \epsilon_{\theta_g}(x_t, C_t, C_c, t) \right\|^2,
\end{aligned}
\end{equation}
where $\epsilon$ is the added noise and $\epsilon_{\theta_g}$ is the prediction from the global appearance model at timestep $t$.

\noindent\textbf{Stage II.}  
The local enhancement model $\theta_l$ refines the coarse latent by injecting logos at user-defined locations. 
To reduce overhead, we fine-tune only the self-attention layers of the logo UNet. Given logo feature $C_l$, spatial mask $C_m$, and garment latent $C_g$, the training objective $L_{\text{logo}}$ is:
\begin{equation}
    L_{\text{logo}} = \mathbb{E}_{x_0, \epsilon, C_l, C_m, C_g, t} \left\| \epsilon - \epsilon_{\theta_l}(x_t, C_l, C_m, C_g, t) \right\|^2,
\end{equation}
where $\epsilon_{\theta_l}$ denotes the prediction from the local enhancement model.

\noindent\textbf{Inference.}  
IMAGGarment supports end-to-end inference through a two-stage pipeline operating in a shared latent space. The global appearance model first generates a latent of coarse garment image conditioned on the input text prompt, silhouette, color, and mask. This process is guided by classifier-free guidance (CFG)~\cite{ho2022classifier}:
\begin{equation}\label{eq:infer_s1}
\begin{split}
\breve{\epsilon}_{\theta_g}(x_t, C_t, C_s, C_c, t)
&= w \cdot \epsilon_{\theta_g}(x_t, C_t, C_s, C_c, t)\\
&\quad + (1 - w) \cdot \epsilon_{\theta_g}(x_t, t)
\end{split}
\end{equation}
here, $w$ is the CFG scale and $x_t$ denotes the noisy latent at timestep $t$.
The coarse latent is then refined by the local enhancement model, which incorporates user-defined logos and spatial constraints through the $A^3$ module. We apply conditional CFG:
\begin{equation}\label{eq:infer_s2}
\begin{split}
\breve{\epsilon}_{\theta_l}(x_t, C_l, C_m, C_g, t)
&= w \cdot \epsilon_{\theta_l}(x_t, C_l, C_m, C_g, t)\\
&\quad + (1 - w) \cdot \epsilon_{\theta_l}(x_t, C_m, C_g, t)
\end{split}
\end{equation}

\section{Experiments}\label{sec:exp}

\subsection{Dataset and Metrics}

\noindent\textbf{Dataset Construction.}
As shown in Fig.~\ref{fig:dataset_pipeline} (a), we construct and release GarmentBench, a large-scale dataset for fine-grained garment generation, containing multi-modal design conditions such as text, sketches, colors, logos, and location masks. It serves as a controllable and extensible benchmark for advancing personalized fashion generation. The construction process is as follows:

\noindent\textbf{\emph{(1) Image Collection and Preprocessing.}}
We collect over 189K high-quality garment images from the internet, covering a wide range of categories such as tops, bottoms, and dresses. To eliminate background distractions and focus on the garment region, we apply YOLOv8~\cite{hussain2024yolov5yolov8yolov10goto} for clothing detection and perform tight cropping to obtain clean garment-centric images for further processing.

\noindent\textbf{\emph{(2) Text, Sketch, and Color Extraction.}}
For each image, we automatically generate three auxiliary conditions to simulate real-world design guidance: textual descriptions generated by the multi-modal LLM Qwen-VL-Chat~\cite{bai2023qwenvlversatilevisionlanguagemodel}, covering key attributes such as color, silhouette, and style; structural sketches obtained using Informative-Drawings~\cite{chan2022learning}, providing shape and layout priors; and color palettes extracted from single-color garments identified via ResNet50~\cite{he2015deepresiduallearningimage} and clustered using K-means~\cite{macqueen1967kmeans}.

\noindent\textbf{\emph{(3) Logo Extraction and Location Annotation.}}
To support logo insertion and spatial control, we further extract local design elements such as logos and prints. We use YOLOv8 to detect visually distinct regions (\emph{e.g.}, anime characters, animal patterns), followed by manual verification to ensure label quality. We also annotate spatial locations and generate binary masks to serve as precise spatial constraints. In total, GarmentBench contains 189,966 garment-condition pairs with rich fine-grained annotations.
    
\noindent\textbf{Dataset Description.} As shown in Fig.~\ref{fig:dataset_pipeline} (b), we present representative samples from the GarmentBench dataset, which include fine-grained garment images paired with multi-modal conditions such as textual descriptions, structural silhouettes, color references, logos, and spatial location masks.
Additionally, we randomly sample images from the Fashion-ControlNet-Dataset-V3\footnote{\url{https://huggingface.co/datasets/Abrumu/Fashion_controlnet_dataset_V3}} and apply the same preprocessing pipeline as GarmentBench to construct a test set with 1,267 image-condition pairs for evaluation and comparative analysis.

\noindent\textbf{Dataset Statement.}
GarmentBench is curated from publicly available fashion imagery under a non-commercial research intent. All personal identifiers were removed; third-party logos and brand marks are included solely to evaluate controllability and remain the property of their respective owners. We release only derived annotations and source URLs (not raw images), together with license notices and a takedown procedure; exact split indices and random seeds are provided for reproducibility.

\noindent\textbf{Evaluation Metrics.}  
We adopt four metrics to comprehensively evaluate visual quality, conditional consistency, and fine-grained controllability. 
Fr\'echet inception distance (FID)~\cite{fid} measures the distribution similarity between generated and real images, reflecting overall realism. 
Color structure similarity (CSS)~\cite{zeng2014color} assesses the consistency of color distribution, measuring color controllability. 
Lastly, Logo location accuracy (LLA)~\cite{fujitake2024rl} quantifies the spatial deviation between generated and target logo positions, reflecting spatial precision.
Learned perceptual image patch similarity (LPIPS)~\cite{lpips} reflects human-perceived visual similarity, effectively capturing structural and textural consistency.
 These metrics comprehensively assess quality and controllability in fine-grained  garment generation.

\subsection{Implementation Details}
In our experiments, both the silhouette UNet and the denoising UNet are initialized with the pretrained Stable Diffusion v1.5 model\footnote{\url{https://huggingface.co/stable-diffusion-v1-5/stable-diffusion-v1-5}}.
The local enhancement model is based on the inpainting variant of Stable Diffusion v1.5\footnote{\url{https://huggingface.co/stable-diffusion-v1-5/stable-diffusion-inpainting}}, with only the self-attention layers being fine-tuned to reduce computational cost.
We adopt OpenCLIP ViT-H/14\footnote{\url{https://github.com/mlfoundations/open_clip}} as the CLIP image encoder. 
All input images are resized to $512\times640$ resolution.
We use the AdamW optimizer~\cite{loshchilov2019decoupledweightdecayregularization} with a constant learning rate of $1\times10^{-5}$. 
The global appearance model and the local enhancement model are trained for 150K and 50K steps, respectively, using a batch size of 20.
During inference, we adopt the DDIM sampler~\cite{song2022denoisingdiffusionimplicitmodels} with 50 sampling steps. 
Unless otherwise specified, the silhouette weight $\alpha$ and color weight $\beta$ in Eq.\ref{eq:silhouette} and Eq.\ref{eq:color} are set to 0.6 and 1.0. 
The classifier-free guidance (CFG) scale $w$ in Eq.\ref{eq:infer_s1} and Eq.\ref{eq:infer_s2} is set to a default value of 7.0.

\begin{table}[t]
  \caption{Quantitative comparisons on GarmentBench. Ours achieves the top results across all metrics, with best in bold.}
  \label{tab:baseline_comparsion}
  \centering
  \scalebox{1.0}{
  \begin{tabular*}{\linewidth}{@{\extracolsep{\fill}}lccccc}
    \toprule
    \textbf{Method} & \textbf{FID} $\downarrow$ & \textbf{CSS} $\downarrow$ & \textbf{LLA} $\uparrow$  & \textbf{LPIPS} $\downarrow$  \\
    \midrule
    BLIP-Diffusion*~\cite{li2023blip} & 101.99 & 104.44 & 0.13 & 0.68 \\
    ControlNet-Garment*~\cite{zhang2023adding} & 41.22 & 83.30 & 0.36 & 0.41 \\
    AnyDoor*~\cite{chen2024anydoor} & 38.08 & 68.24 & 0.65 & 0.17  \\
    IP-Adapter-Garment*~\cite{ye2023ip} & 37.95 & 92.95 & 0.36 & 0.43  \\
    \midrule
    \textbf{IMAGGarment (Ours)} & \textbf{17.63} & \textbf{36.16} & \textbf{0.72} & \textbf{0.10} \\
  \bottomrule
\end{tabular*}
}
\hfill\parbox{0.95\linewidth}{\footnotesize\textit{* denotes re-implemented by us for a fair comparison.}}
\end{table}

\subsection{Baseline Comparisons}
\begin{figure*}[t]
    \centering
    \includegraphics[width=\linewidth]{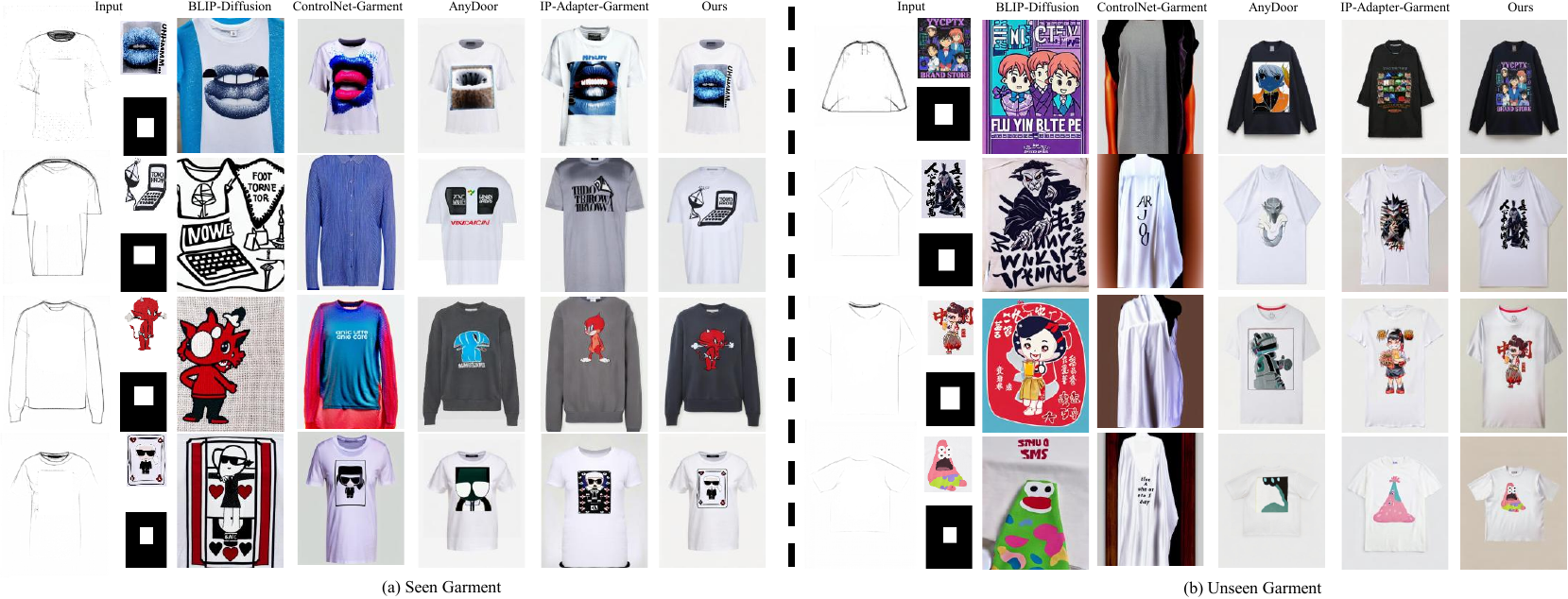}
    \vspace{-0.5cm}
    \caption{\textbf{Qualitative results on seen and unseen GarmentBench samples.} The seen set uses original test pairs, while the unseen set involves randomly mixed conditions. IMAGGarment delivers the most consistent outputs, achieving accurate silhouette, color, and logo control across both settings.}
    \vspace{-0.3cm}
    \label{fig:comparsion}
\end{figure*}

Due to the absence of prior work tailored to fine-grained garment generation with multi-condition control, we compare our method against four representative baselines: BLIP-Diffusion~\cite{li2023blip}, AnyDoor~\cite{chen2024anydoor}, ControlNet~\cite{zhang2023adding}, and IP-Adapter~\cite{ye2023ip}. 
For subject-driven generation methods, BLIP-Diffusion~\cite{li2023blip} leverages a learnable Q-Former to align textual and visual embeddings in the latent space, initially designed for subject-preserving generation from text-image pairs. 
AnyDoor~\cite{chen2024anydoor} combines identity and detail encoders to reconstruct personalized content, which we adapt to conditions of garment appearance and logo inputs. 
For plugin-based baselines, we extend ControlNet~\cite{zhang2023adding} and IP-Adapter~\cite{ye2023ip} by duplicating and modifying their conditional branches to support multi-conditional inputs, such as silhouette, color, and logo. 
The adapted versions are referred to as ControlNet-Garment and IP-Adapter-Garment. Specifically, for ControlNet-Garment, we input silhouette, color, logo and mask maps into the ControlNet branch and inject them at each downsampling block, following standard practice. For IP-Adapter-Garment, we extend the official implementation to accept silhouette, color, logo and mask embeddings, which are concatenated and injected via cross-attention. 
To ensure task relevance, all methods are fine-tuned on our GarmentBench dataset with support for logo-specific conditioning. 
All methods are trained and evaluated under identical training protocols, input resolutions, and hardware setups.
The corresponding quantitative and qualitative results are presented in Table~\ref{tab:baseline_comparsion} and Fig.~\ref{fig:comparsion}, respectively, with detailed analysis provided below.

\begin{table}[t]
  \caption{Quantitative ablation results on GarmentBench.}
  \label{tab:ablation}
  \vspace{-.1cm}
  \scalebox{1.0}{
  \begin{tabular*}{\linewidth}{@{\extracolsep{\fill}}lccccc}
    \toprule
    \textbf{Method} & \textbf{FID} $\downarrow$ & \textbf{CSS} $\downarrow$ & \textbf{LLA} $\uparrow$  & \textbf{LPIPS} $\downarrow$ \\
    \midrule
    B0 & 139.33& 104.54 & 0.15 & 0.64 \\
    B1 & 47.42& 36.65& 0.30& 0.15  \\
   B2 & 30.19& 97.05& 0.56 & 0.33\\
   B3 & 21.20& 43.00& 0.65& 0.11 \\
   B4 & 46.16& 108.25& 0.52& 0.38  \\
    \midrule
    \textbf{Full} &  \textbf{17.63} & \textbf{36.16} & \textbf{0.72} & \textbf{0.10} \\
  \bottomrule
\end{tabular*}
}
\end{table}

\begin{figure*}[t]
    \centering
\includegraphics[width=0.95\linewidth]{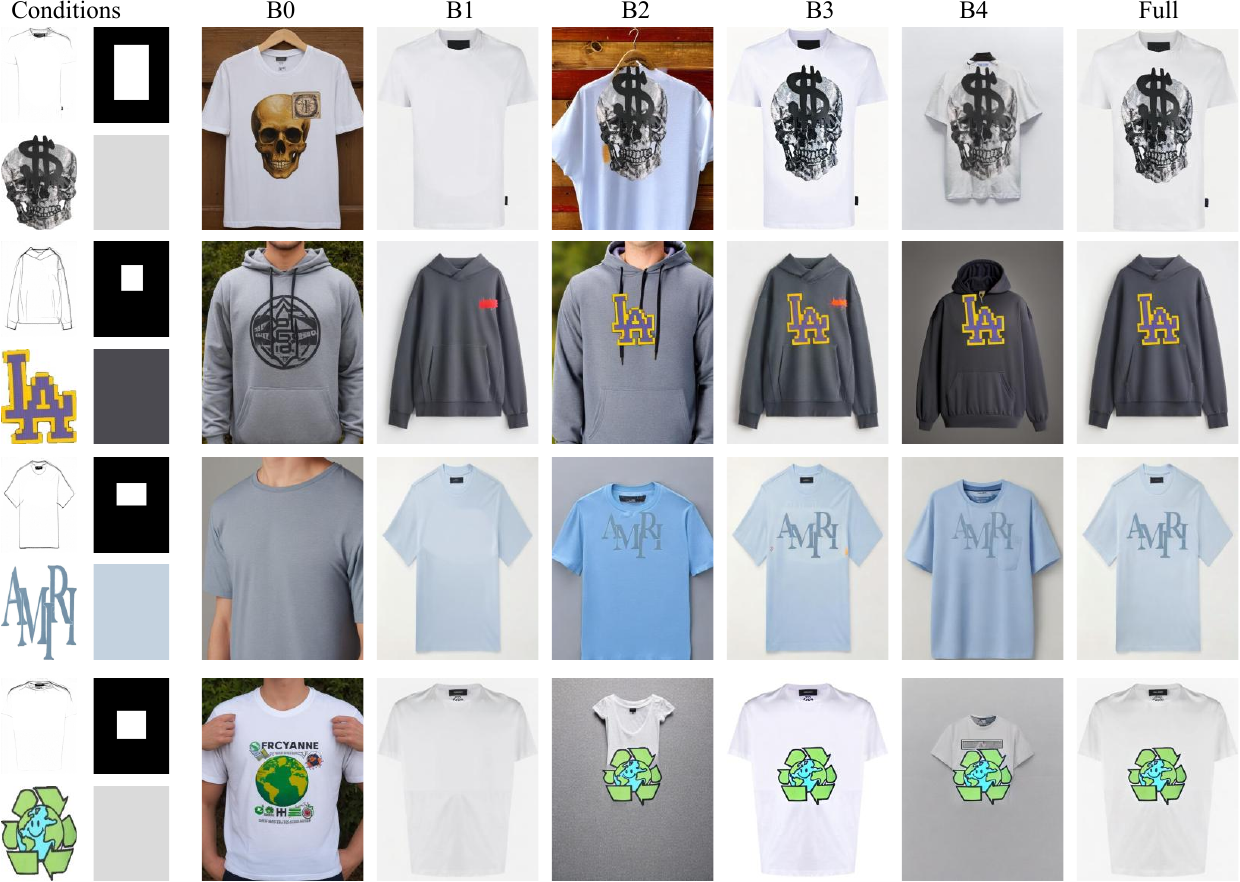}
    \vspace{-0.3cm}
    \caption{\textbf{Qualitative ablation results on GarmentBench.} The “Full” configuration achieves the best results, highlighting the
importance of each component.}
    \vspace{-0.3cm}
\label{fig:ablation_result}
\end{figure*}

\noindent\textbf{Quantitative Results.} 
As shown in Table~\ref{tab:baseline_comparsion}, IMAGGarment achieves the best performance across all four metrics on the GarmentBench dataset, demonstrating its superiority in controllable fine-grained garment generation.
Compared to subject-driven methods (BLIP-Diffusion~\cite{li2023blip}, AnyDoor~\cite{chen2024anydoor}), which rely on global features for personalized reconstruction, IMAGGarment shows substantial improvements in {FID}, {LPIPS}, and {CSS}. These gains highlight the effectiveness of our mixed attention and color adapter modules in achieving coherent multi-condition fusion, resulting in more realistic, perceptually consistent, and color-faithful outputs.
In contrast to plugin-based approaches (ControlNet-Garment~\cite{zhang2023adding}, IP-Adapter-Garment~\cite{ye2023ip}) that simply stack independent conditional branches, IMAGGarment yields significantly higher {LLA}, reflecting more precise logo placement. 
Our proposed A$^3$ module drives these improvements, which adaptively injects spatial priors and logo features into the latent space for accurate local control.
Overall, these results indicate that global-only conditioning or naive plugin stacking is insufficient for fine-grained control. 
By contrast, IMAGGarment provides an effective solution for multi-conditional garment synthesis, enabling precise coordination of global structure and local detail.

\noindent\textbf{Qualitative Results.}
Fig.~\ref{fig:comparsion} presents qualitative comparisons on both seen and unseen garments. 
Notably, the seen test set refers to the designated test split of our GarmentBench dataset. 
In the absence of other suitable public datasets, we assess generalization using an unseen-composition test split constructed by randomly recombining input conditions (\emph{e.g.}, silhouette, color, logo) into combinations that never appear during training, thereby simulating real-world fashion-design scenarios.
On seen garments, subject-driven methods (BLIP-Diffusion~\cite{li2023blip}, AnyDoor~\cite{chen2024anydoor}) reconstruct global appearance but lack spatial control. BLIP-Diffusion retains logo identity yet fails at precise placement due to text-only conditioning, while AnyDoor introduces logo distortions and stylistic artifacts. Plugin-based baselines (ControlNet-Garment~\cite{zhang2023adding}, IP-Adapter-Garment~\cite{ye2023ip}) treat conditions independently, resulting in poor coordination. ControlNet-Garment suffers from cross-condition interference, and IP-Adapter-Garment often misplaces logos despite preserving structure.
In contrast, IMAGGarment achieves accurate control over silhouette, color, and logo placement.
On unseen garments, all baselines degrade notably. 
Subject-driven methods fail to generalize to novel layouts, AnyDoor distorts appearance, and BLIP-Diffusion struggles with logo positioning. Plugin-based methods also falter: ControlNet-Garment produces mismatched outputs, and IP-Adapter-Garment cannot interpret unseen spatial semantics.
IMAGGarment remains robust, maintaining alignment across all conditions. This generalization stems from our $A^3$ module, which effectively integrates spatial and visual cues in the latent space. These results validate the controllability and flexibility of our method in both seen and unseen settings.

\begin{figure*}[t]
    \centering
    \includegraphics[width=0.95\linewidth]{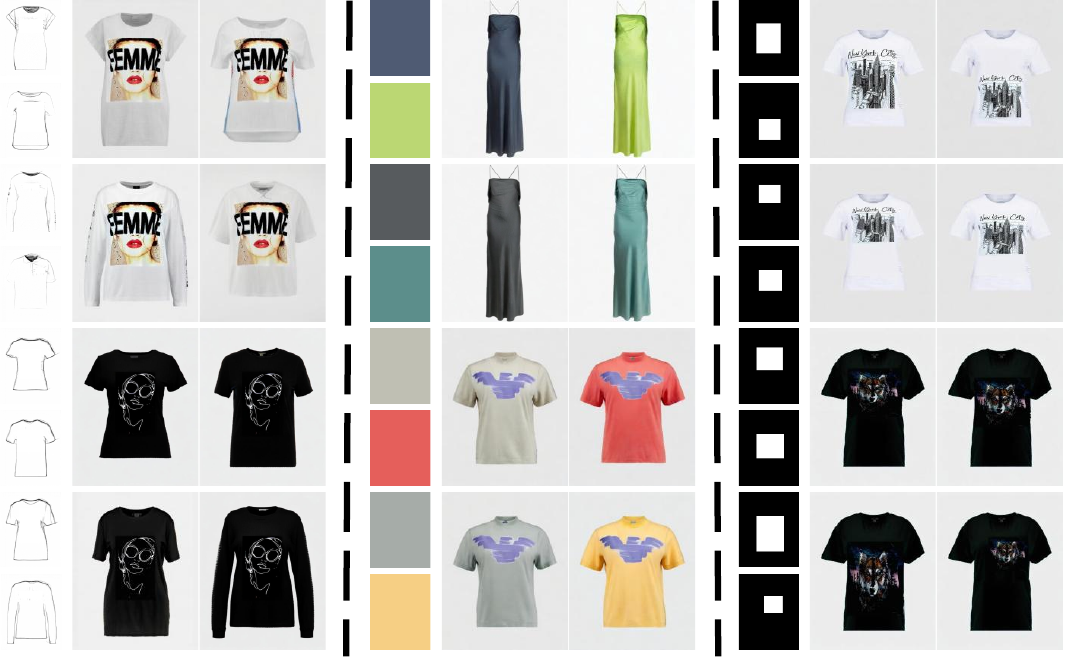}
    \vspace{-0.5cm}
    \caption{\textbf{Controllability visualization.} Each block varies one input condition while keeping others fixed. Left: Silhouette changes lead to consistent structural adaptation. Middle: Color palette variation results in accurate color transfer. Right: Logo mask adjustment yields precise spatial placement.}
       \vspace{-0.3cm}
    \label{fig:controllability}
\end{figure*}

\subsection{Ablation Study}
 
To validate the effectiveness of each component in our framework, we design a series of ablation variants within the IMAGGarment architecture:
\textbf{B0} uses the vanilla Stable Diffusion v1.5 without any of our proposed modules, serving as the baseline.
\textbf{B1} removes the local enhancement model (Stage II), evaluating the impact of omitting logo injection and spatial control.
\textbf{B2} removes the global appearance model (Stage I), assessing the model's performance without structured silhouette and color conditioning.
\textbf{B3} removes the color adapter from the global appearance model, isolating the role of color guidance in generation.
\textbf{B4} replaces our mixed attention with vanilla self-attention in the denoising UNet, testing the importance of spatial fusion with silhouette features.
\textbf{Full} represents the complete IMAGGarment framework with all proposed modules integrated.

\noindent\textbf{Ablation of Architecture Design.} Table~\ref{tab:ablation} presents the quantitative impact of each component in our proposed IMAGGarment.
In B1, which removes the local enhancement stage, the model struggles to place logos precisely, leading to degraded LLA. Although the overall garment structure is preserved, the lack of spatial control prevents accurate logo integration.
In B2, without the global appearance stage, the model fails to maintain silhouette and color consistency, resulting in significantly worse FID, LPIPS, and CSS. This demonstrates that local injection alone is insufficient to handle global garment layouts.
B3 disables the color adapter, causing notable drops in CSS, highlighting its role in faithful color transfer and control.
B4 replaces our mixed attention with standard self-attention, which weakens the fusion of silhouette guidance and causes drops in both LPIPS and FID, indicating reduced realism and structural coherence.
The full IMAGGarment achieves the best performance across all metrics, validating the complementary design of each module's effectiveness in handling multi-condition garment generation.
Further, Fig.~\ref{fig:ablation_result} shows qualitative comparisons.
B1 fails to align logos spatially, while B2 produces distorted garments lacking color and silhouette guidance. Despite maintaining logo placement, B3 leads to color mismatch, and B4 generates less coherent garment layouts. In contrast, the full model successfully synthesizes garments with accurate silhouettes, precise logo placement, and faithful color reproduction, demonstrating the benefits of our dual-stage design, color adapter, and mixed attention fusion.
Overall,  The “Full” configuration achieves the best results, highlighting the
importance of each component.

\begin{figure*}[t]
    \centering
    \includegraphics[width=0.95\linewidth]{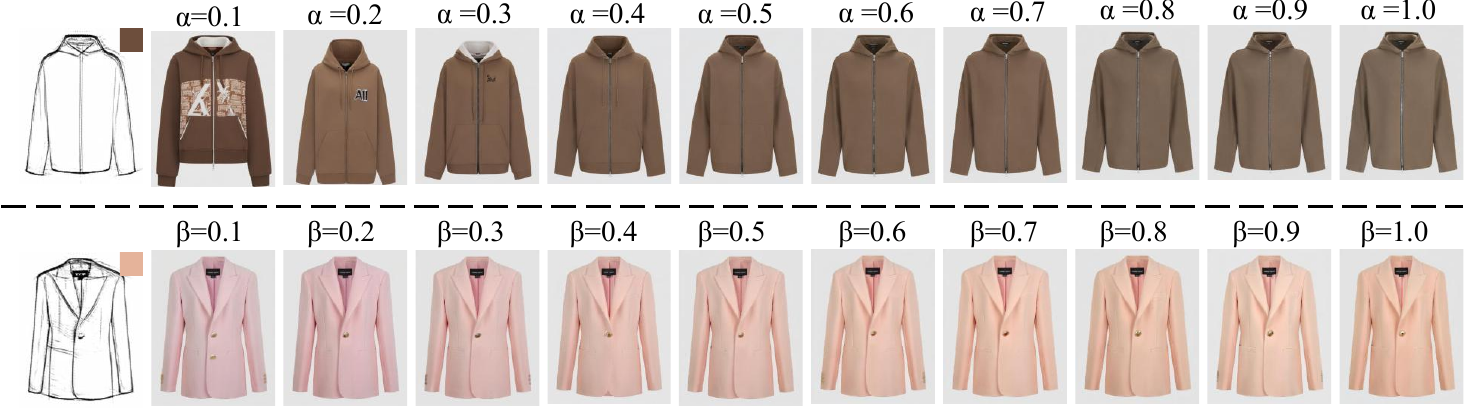}
    \vspace{-0.3cm}
    \caption{\textbf{Hyperparameter analysis of silhouette weight $\alpha$ and color weight $\beta$.}}
     \vspace{-0.3cm}
    \label{fig:hyperparam}
\end{figure*}

\subsection{More Results and Analysis}

\noindent\textbf{Controllability Analysis.}
We assess controllability by varying a single condition at a time (silhouette, color palette, or logo position) while keeping the others fixed. As shown in Fig.~\ref{fig:controllability}, each three column block visualizes the model's response to one condition.
Changing the silhouette (left block) yields garments that match the target shapes, indicating that the mixed attention module preserves structural alignment. Varying the color palette (middle block) produces the intended color distributions, validating the color adapter for color faithful generation. Adjusting the logo position (right block) achieves precise spatial relocation, showing that the $A^3$ module effectively injects spatial priors for local control.
Overall, IMAGGarment provides fine-grained and decoupled control of garment attributes suitable for practical design workflows. Non-varied attributes remain stable across manipulations, reflecting minimal cross-condition interference and consistent editing behavior. Sequential composition of edits across attributes produces similar outcomes regardless of edit order, which suggests low inter-attribute coupling. Control fidelity also holds under moderate changes of viewpoint and background, supporting robustness in real design scenarios.

\noindent\textbf{Hyperparameter Analysis.}
We study the effect of two key hyperparameters in Eq.\ref{eq:silhouette} and Eq.\ref{eq:color}: the silhouette guidance weight $\alpha$ and the color conditioning weight $\beta$. From Fig.~\ref{fig:hyperparam}, varying $\alpha$ directly impacts the model's ability to follow the reference silhouette. When $\alpha$ is too low, the generated structure becomes blurry or deviates from the target shape; when too high, it may suppress color and text guidance. We empirically set $\alpha=0.6$ for balanced structural alignment. Similarly, the color weight $\beta$ controls the influence of the color palette. As $\beta$ increases, color consistency improves steadily, with $\beta=1.0$ yielding the best visual fidelity. Joint sweeps over $(\alpha,\beta)$ indicate a broad stability region around $\alpha\in[0.5,0.7]$ and $\beta\in[0.8,1.1]$, showing robustness to moderate mistuning. Interaction effects are mild: very large $\alpha$ slightly narrows the effective range of $\beta$, while very large $\beta$ can oversaturate colors and reduce shading nuance. We therefore adopt $\alpha=0.6$ and $\beta=1.0$ throughout all experiments.


\section{Conclusion}\label{sec:conclusion}
We propose IMAGGarment, a unified conditional diffusion framework for fine-grained garment generation with precise control over silhouette, color, and logo placement. By introducing mixed attention, color adapter, and the $A^3$ module, our framework explicitly disentangles global structure (silhouette and color) from local attributes (logo content and spatial placement), enabling accurate spatial control and high-quality synthesis.
To support this task, we construct GarmentBench, a large-scale benchmark with over 180K samples annotated with multi-level design conditions.
Comprehensive experiments on both seen and unseen garments demonstrate that IMAGGarment achieves state-of-the-art results in structure fidelity, color consistency, and logo controllability. Code, models, and datasets are publicly available at \url{https://github.com/muzishen/IMAGGarment}.

\bibliographystyle{unsrt}
\bibliography{ref}

\end{document}